%% file: main.tex
\documentclass[letterpaper,conference]{IEEEtran}

\newif\ifarxiv
\arxivtrue

\IEEEoverridecommandlockouts %
\usepackage[utf8]{inputenc}
\usepackage{cite}
\usepackage{amsmath,amssymb,amsfonts}
\usepackage{algorithmic}
\usepackage{graphicx}
\usepackage{textcomp}
\usepackage[dvipsnames,table]{xcolor}
\usepackage{amsmath}
\usepackage{amssymb}
\usepackage{tabularx}
\usepackage{enumitem}
\usepackage{url}
\usepackage{tikz}
\usepackage{booktabs}
\usepackage{anyfontsize}
\usepackage{pgfplots}
\pgfplotsset{compat=1.18}
\usepackage{multirow}
\usepackage[export]{adjustbox}
\usepackage[para,online,flushleft]{threeparttable}
\usepackage{booktabs}
\usepackage[table]{xcolor}
\usepackage[nolist]{acronym}
\usepackage{makecell}
\usepackage[caption=false]{subfig}
\usepackage{svg}
\usepackage{xcolor}

\usetikzlibrary{patterns,patterns.meta}
\usetikzlibrary{shadows,arrows.meta,positioning,calc}
\def\BibTeX{{\rm B\kern-.05em{\sc i\kern-.025em b}\kern-.08em
    T\kern-.1667em\lower.7ex\hbox{E}\kern-.125emX}}

\newif\ifieee
\ieeefalse

\ifieee
\else
\usepackage[colorlinks=true,allcolors=black]{hyperref} %
\fi

\usepackage[capitalise,noabbrev]{cleveref} %
\usepackage{makecell}

\newcommand*{\RL}[2][]{\textcolor{Rhodamine}{[\textbf{\ifthenelse{\equal{#1}{}}{RL}{RL(#1)}}: #2]}}

\begin{acronym}[TDMA]
    \acro{SGD}{Stochastic Gradient Descent}
	\acro{SVM}{Support Vector Machine}
\end{acronym}

\usepackage{transparent}
\usepackage{tikz}
\ifarxiv
    \newcommand\copyrighttext{%
      \scriptsize Accepted at SMC 2026. The final published version will be available in IEEE Xplore.}
    \newcommand\copyrightnotice{%
    \begin{tikzpicture}[remember picture,overlay]
    \node[anchor=south,yshift=30pt,xshift=0pt] at (current page.south) {\fbox{\transparent{0.85}\parbox{\dimexpr0.48\textwidth-\fboxsep-\fboxrule\relax}{\copyrighttext}}};
    \end{tikzpicture}%
    }
\else
\fi

\begin{document}

\title{Toward Parking Spot Occupancy Recognition:\\A Self-Supervised Approach\\ \thanks{This work has been supported by the Brazilian National Council for Scientific and Technological Development (CNPq) – Grant 405511/2022-1.}} %

\author{
\IEEEauthorblockN{
Luan Marko Kujavski\IEEEauthorrefmark{1},
Rayson Laroca\IEEEauthorrefmark{2}\textsuperscript{,}\IEEEauthorrefmark{1},
Paulo Lisboa de Almeida\IEEEauthorrefmark{1}
}
\IEEEauthorblockA{
\IEEEauthorrefmark{1}\hspace{0.15mm}Federal University of Paraná, Curitiba, Brazil
}
\IEEEauthorblockA{
\IEEEauthorrefmark{2}\hspace{0.15mm}Pontifical Catholic University of Paraná, Curitiba, Brazil
}
\IEEEauthorrefmark{1}{\hspace{-0.2mm}\tt\small\{luan.marko, paulorla\}@ufpr.br} 
\quad 
\IEEEauthorrefmark{2}{\hspace{0.15mm}\tt\small \texttt{rayson@ppgia.pucpr.br}}
}

\maketitle

\ifarxiv
    \copyrightnotice
\else
\fi

\input{abstract_keywords}

\input{1-introduction/introduction}
\input{2-related_works/related_works}
\input{3-training_pipeline/training_pipeline}
\input{4-experiment_protocol/experiment_protocol}
\input{5-results/results}
\input{6-conclusion/conclusion}

\bibliographystyle{IEEEtran}
\bibliography{references}

\end{document}

%% file: abstract_keywords.tex
\begin{abstract}
As urban areas expand, automatic monitoring of parking lots becomes essential for efficient and sustainable cities. This work proposes a self-supervised approach for parking spot occupancy recognition that requires no labeled samples from the target parking lot. Building upon a self-supervised transfer learning fine-tuning protocol, the proposed training strategy consists of two self-supervised stages: first on unlabeled generic data and then on unlabeled target-specific data, followed by supervised fine-tuning using only generic parking lot labels. We adopt SimCLR with a ResNet-50 encoder and evaluate the method under a leave-one-out cross-environment protocol on three public datasets: PKLot, CNRPark-EXT, and PLds. We also introduce a two-stage deployment strategy in which a \emph{Strong General Model} is initially deployed, followed by a \emph{Specialized Model} that incorporates unlabeled images collected during the first $N$ days of deployment in a self-supervised manner. Experimental results show that the \emph{Strong General Model} alone outperforms supervised and self-supervised baselines, achieving an average accuracy of 97.2\%, which further improves to 97.8\% with the proposed two-stage strategy. These results demonstrate that self-supervised learning enables a scalable and label-efficient solution for real-world parking occupancy monitoring. Our trained models and source code are publicly available at \url{https://github.com/LoanMaikon/Parking-Spot-Occupancy-Recognition}.

\end{abstract}


%% file: 1-introduction/introduction.tex
\section{Introduction}

As urban areas continue to expand in both size and population density, the time drivers spend searching for available parking spots contributes substantially to traffic congestion, increased fuel consumption, and higher carbon emissions~\cite{emissao_gases}. 
In this context, vision-based automated parking spot occupancy classification (which determines whether a parking space is empty or occupied) emerges as a key enabling technology for more efficient and sustainable urban environments. By enabling real-time monitoring of parking facilities, such systems support faster, data-driven decision-making for both drivers and city management platforms.

Despite recent advances~\cite{pklot,cnrpark,plds,paulo_review,hochuli,hochuli2022annotStrat,paulo_distill}, the high human cost associated with labeling data from the target environment remains a major challenge~\cite{paulo_review}. In the absence of such labeled data, most models suffer a significant drop in accuracy when deployed in unseen environments~\cite{paulo_review,hochuli}. In this work, we propose a self-supervised learning approach, coupled with a dedicated training pipeline, to create robust general and specialized models that do not rely on labeled data from the target environment. To the best of our knowledge, self-supervised learning remains an underexplored family of methods in the context of vision-based parking spot occupancy recognition.

\begin{figure}[!htbp]
    \centering
    \includegraphics[width=1\linewidth]{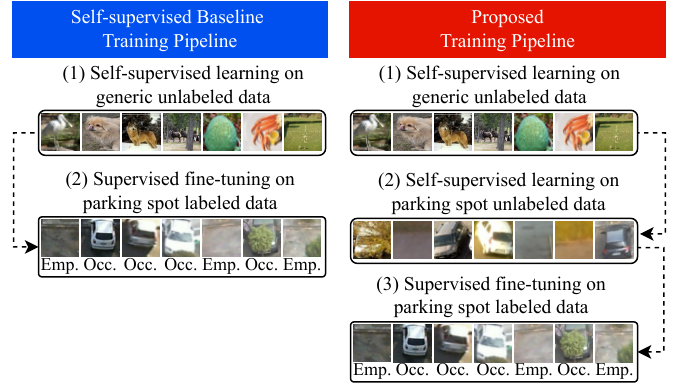}

    \vspace{-1mm}
    
    \caption{Overview of the proposed training pipeline (right), which comprises two self-supervised stages followed by supervised fine-tuning. This pipeline extends the standard self-supervised transfer learning protocol (left).}
    \label{fig:pipeline}
\end{figure}

A typical self-supervised evaluation under a transfer learning via fine-tuning protocol consists of a pretraining stage on generic unlabeled data, followed by supervised fine-tuning on task-specific data~\cite{simclr}. We refer to this standard approach as the \emph{Self-supervised Baseline}. Inspired by~\cite{medical_image_classification}, we extend this protocol by introducing an additional self-supervised fine-tuning stage using domain-specific data (i.e., parking lot images). As illustrated in \cref{fig:pipeline}, the proposed training scheme follows a three-stage pipeline: (1) self-supervised pretraining on generic data (ImageNet~\cite{imagenet}); (2) self-supervised fine-tuning on parking lot data; and (3) supervised fine-tuning using labeled samples from generic parking lot data.

The pipeline shown on the right in \cref{fig:pipeline} yields a \emph{Strong General Model} for generic parking spot classification. Inspired by \cite{paulo_distill}, we further propose and evaluate a two-stage deployment scheme, illustrated in \cref{fig:deploy_scheme}. In the first stage, the \emph{Strong General Model} is deployed in the first $N$ days. After the $N$-th day, a \emph{Specialized Model} is trained using the same pipeline depicted in \cref{fig:pipeline}, incorporating unlabeled samples collected from the target parking lot during the first $N$ days together with the generic parking lot data in stage~(2).

We evaluate our approach on three public parking spot occupancy recognition datasets: PKLot~\cite{pklot}, CNRPark-EXT~\cite{cnrpark}, and PLds~\cite{plds}. These datasets cover a broad range of real-world conditions, including variations in lighting, weather, and viewpoints~\cite{paulo_review}. The main findings of this work are summarized in \cref{fig:summary}. The proposed \emph{Two-stage Deployment} scheme achieves the highest accuracy, followed by the \emph{Strong General Model}. We compare our approach against two baselines: the \emph{Supervised Baseline} and the \emph{Self-supervised Baseline}, which follow standard transfer learning via fine-tuning protocols~\cite{simclr}. These baselines consist of a pretraining stage on generic data, performed either in a supervised or self-supervised manner, followed by supervised fine-tuning on task-specific data.

\definecolor{c1}{HTML}{2ECC71}
\definecolor{c2}{HTML}{9B59B6}
\definecolor{c3}{HTML}{3498DB}
\definecolor{c4}{HTML}{E74C3C}

\begin{figure}[htbp]
    \centering

    \begin{tabular}{m{0.45\linewidth} m{0.45\linewidth}}
    \tikz{
    \fill[c1] (0,0) rectangle (0.18,0.18);
    \fill[pattern=north east lines, pattern color=black] (0,0) rectangle (0.18,0.18);
    }~\footnotesize Self-supervised Baseline &
    \tikz{
    \fill[c3] (0,0) rectangle (0.18,0.18);
    \fill[pattern=north west lines, pattern color=black] (0,0) rectangle (0.18,0.18);
    }~\footnotesize Strong General Model \\
    \tikz{
    \fill[c2] (0,0) rectangle (0.18,0.18);
    \fill[pattern=vertical lines, pattern color=black] (0,0) rectangle (0.18,0.18);
    }~\footnotesize Supervised Baseline &
    \tikz{
    \fill[c4] (0,0) rectangle (0.18,0.18);
    \fill[pattern=horizontal lines, pattern color=black] (0,0) rectangle (0.18,0.18);
    }~\footnotesize Two-stage Deployment
    \end{tabular}

    \vspace{1mm}
    \input{1-introduction/tikz/summary}
    
    \vspace{-3mm}
    
    \caption{Summary of the main findings of this work. The reported results correspond to the average accuracy across all experiments.}
    \label{fig:summary}
\end{figure}
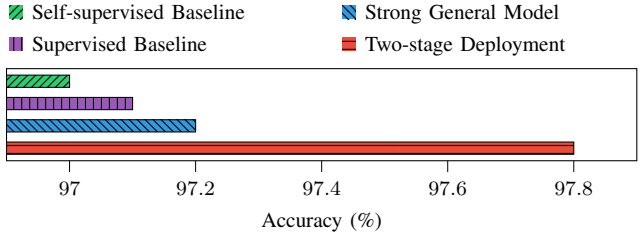

Our results show that the proposed approach achieves performance comparable to methods that rely on labeled samples, while eliminating the need for labeled data from the target deployment environment, which remains a major bottleneck in parking spot occupancy research~\cite{paulo_review}. Accordingly, the main contributions of this work are:

\begin{itemize}
    \item We propose a self-supervised learning-based training pipeline to obtain a strong parking spot occupancy classifier without relying on labeled data from the target.
    \item We show that unlabeled data collected from the target parking lot can be effectively exploited to train a specialized model tailored to the deployment scenario.
\end{itemize}

The remainder of this paper is organized as follows. \cref{sec:related_works} reviews the related literature. The proposed approach is described in \cref{sec:approach}. The experimental protocol and results are presented and discussed in \cref{sec:experiment_protocol,sec:results}, respectively. Finally, conclusions are drawn in \cref{sec:conclusion}.

%% file: 1-introduction/tikz/summary.tex
\begin{tikzpicture}

\definecolor{darkgray176}{RGB}{176,176,176}
\definecolor{dodgerblue52152219}{RGB}{52,152,219}
\definecolor{mediumorchid15589182}{RGB}{155,89,182}
\definecolor{mediumseagreen46204113}{RGB}{46,204,113}
\definecolor{tomato2317660}{RGB}{231,76,60}

\begin{axis}[
font=\footnotesize,
height=2.8cm,
tick align=outside,
width=1.12\columnwidth,
x grid style={darkgray176},
xlabel={Accuracy (\%)},
xmin=96.9, xmax=97.9,
xtick pos=left,
xtick style={color=black},
y dir=reverse,
ymajorticks=false,
ymin=-0.35, ymax=2.21
]
\draw[draw=black,fill=mediumseagreen46204113,postaction={pattern=north east lines}]
(axis cs:96.9,-0.17) rectangle (axis cs:97.0,0.17);
\draw[draw=black,fill=mediumorchid15589182,postaction={pattern=vertical lines}]
(axis cs:96.9,0.45) rectangle (axis cs:97.1,0.79);
\draw[draw=black,fill=dodgerblue52152219,postaction={pattern=north west lines}]
(axis cs:96.9,1.07) rectangle (axis cs:97.2,1.41);
\draw[draw=black,fill=tomato2317660,postaction={pattern=horizontal lines}]
(axis cs:96.9,1.69) rectangle (axis cs:97.8,2.03);
\end{axis}

\end{tikzpicture}

%% file: 2-related_works/related_works.tex
\section{Related Work}
\label{sec:related_works}

This Section reviews prior work on self-supervised learning and parking spot occupancy classification, emphasizing the main paradigms and how they motivate the proposed approach.

\subsection{Self-Supervised Learning}

Self-supervised learning has recently emerged as a powerful paradigm for representation learning without labeled data.
Instead of relying on annotations, these methods employ pretext tasks that enable models to learn invariant features capturing semantic information from large amounts of unlabeled data.
Once trained, the learned representations can be transferred to downstream tasks, typically by adding a task-specific head on top of the learned features~\cite{simclr,moco,swav,dino,ijepa}.

Determining when self-supervised learning was first introduced is challenging.
Nevertheless, early work such as~\cite{virginia_1993} demonstrated that algorithms could learn to classify data without explicit external labels by exploiting relationships across multiple modalities.
Subsequent approaches focused on learning visual representations by associating different regions of an image, for instance by predicting relative patch positions~\cite{context_prediction}, identifying the correct permutation of shuffled patches~\cite{jigsaw}, or performing inpainting-based prediction tasks~\cite{inpainting}.

More recent approaches, as discussed in~\cite{cookbook}, explore strategies such as canonical correlation analysis, in which the model jointly learns feature representations and assigns samples to prototype clusters~\cite{swav}; self-distillation, where a network learns from its own predictions~\cite{dino}; and masked image modeling, which trains the model to reconstruct missing or masked regions of the input image~\cite{ijepa}.
Another line of work, explored in this paper, relies on deep metric learning, where models are trained to pull similar samples closer in the embedding space while pushing dissimilar ones apart~\cite{simclr,moco}.
Inspired by~\cite{medical_image_classification}, we incorporate a deep metric self-supervised approach into the parking spot occupancy classification problem, enabling the exploitation of unlabeled data from the target parking lot and the learning of richer, domain-specific features.

\subsection{Parking Spot Occupancy Classification}

Related work on parking spot occupancy classification can be broadly divided into three phases.
In the first phase, researchers focused on the creation of large-scale datasets for training and evaluation, including PKLot~\cite{pklot}, CNRPark-EXT~\cite{cnrpark}, and PLds~\cite{plds}.

The second phase is characterized by the development of feature extraction and deep learning based methods capable of achieving high accuracy rates, often above~99\%, provided that a large number of labeled samples from the target parking lot are available.
Representative works include~\cite{pklot,cnrpark}, with a comprehensive overview presented in~\cite{paulo_review}.
Despite their strong performance, these approaches are costly, as they require extensive data annotation whenever the system is deployed.

The third and current phase focuses on reducing or eliminating the need for labeled samples from the target environment and on developing lightweight models suitable for deployment on power-restricted platforms, such as edge devices~\cite{paulo_review}.
Representative works include~\cite{pklot,cnrpark,paulo_distill,hochuli,hochuli2022annotStrat}.
The methods proposed in~\cite{pklot} and~\cite{cnrpark} introduced models based on \acp{SVM} and deep learning, respectively.
Although computationally efficient, these models require large amounts of labeled data from the target for training.

The authors in~\cite{hochuli} reduced the labeling effort by showing that, with approximately 1,000 labeled samples from the target environment, a model can be fine-tuned to achieve accuracies close to~97\%.
This gap was reduced in~\cite{paulo_distill}, where similar performance was achieved without using labeled data from the target environment.
In that work, a teacher-student framework was adopted, in which a teacher model generates pseudo-labels during the first $N$ days of deployment, and a lightweight model is subsequently trained using these pseudo-labels.

In this work, we draw inspiration from~\cite{paulo_distill} and~\cite{medical_image_classification} to combine self-supervised learning with the collection of unlabeled data from the target environment.
This strategy enables the fine-tuning of a custom model that can be deployed directly on the target device, such as a smart camera, without requiring labeled data from the target environment.
Unlike the method proposed in~\cite{paulo_distill}, our approach does not rely on a teacher model.

%% file: 3-training_pipeline/training_pipeline.tex
\begin{figure*}[!htbp]
    \centering
    \includegraphics[width=0.925\linewidth]{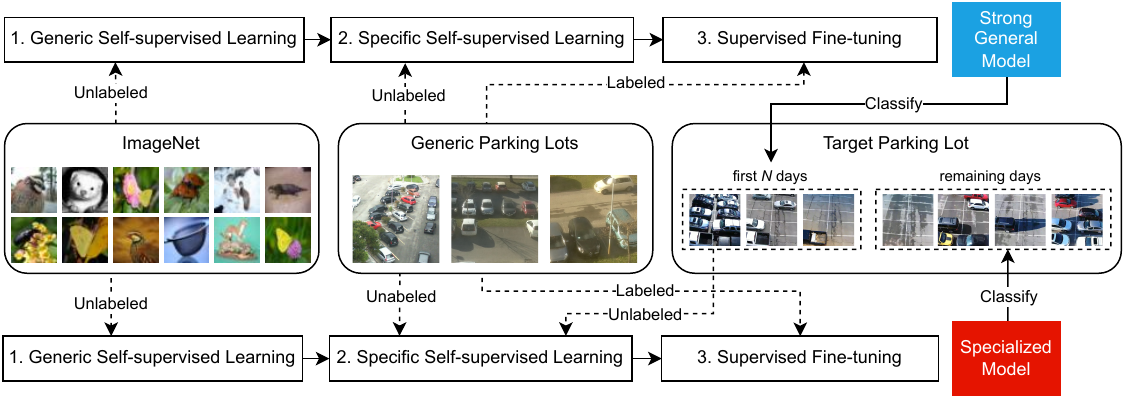}

    \vspace{-1.5mm}
    
    \caption{Proposed two-stage deployment scheme. During the first $N$ days, classification is performed using the \emph{Strong General Model}. Afterward, classification is carried out by the \emph{Specialized Model}, trained by incorporating unlabeled data collected during the first $N$ days into the initial generic parking lot data.}
    \label{fig:deploy_scheme}

    \vspace{-0.35cm}
\end{figure*}

\section{Proposed Approach}
\label{sec:approach}

The proposed approach leverages a self-supervised learning strategy~\cite{simclr,medical_image_classification} to build two models: a \emph{Strong General Model}, designed to classify instances from diverse parking lots, and a \emph{Specialized Model}, tailored to achieve high performance on a specific target parking lot.

As illustrated in \cref{fig:deploy_scheme}, the \emph{Strong General Model}, trained using generic parking lot data, is deployed during the first $N$ days of operation.
After the $N$-th day, a \emph{Specialized Model} is trained by augmenting the initial generic dataset with unlabeled parking spot images collected from the target environment during this initial period.
Both models follow the training pipeline depicted in \cref{fig:pipeline}~(right) and require no labeled data from the target parking lot, eliminating the need for manual annotation during real-world deployment.

By relying on a pre-trained encoder and a self-supervised learning paradigm, the computational cost of training is substantially reduced, since generic visual representations are learned during pretraining.
Both the \emph{Strong General Model} and the \emph{Specialized Model} are initialized from a encoder pre-trained using the SimCLR framework~\cite{simclr}.
SimCLR learns visual representations by maximizing agreement between representations of different augmented views of the same image while contrasting them with representations from other images in the batch.
For the $i$-th image in the batch and its corresponding $j$-th augmented view, the NT-Xent loss is defined as

\begin{equation}
\ell_{i,j} =
- \log
\frac{\exp(\mathrm{sim}(\mathbf{z}_i,\mathbf{z}_j)/\tau)}
{\sum_{k=1}^{2N} \mathbf{1}_{[k\ne i]}\,\exp(\mathrm{sim}(\mathbf{z}_i,\mathbf{z}_k)/\tau)},
\end{equation}

\noindent where $\mathrm{sim}(\cdot, \cdot)$ denotes the cosine similarity between two vectors, $\tau$ is a temperature scalar, $\mathbf{z}$ represents the output of the projection head, $N$ is the batch size, and $k$ indexes the $k$-th view in the batch.
As in~\cite{simclr}, the encoder is initially pre-trained on the ImageNet dataset~\cite{imagenet}.

In the second stage, the encoder is further fine-tuned through an additional self-supervised training stage using domain-specific datasets, allowing it to capture task-relevant characteristics.
A similar strategy was adopted in~\cite{medical_image_classification} for the medical domain.
At this point, the training procedures for the \emph{Strong General Model} and the \emph{Specialized Model} diverge.
The former is fine-tuned using unlabeled data from generic parking lots.
The latter is fine-tuned using the same unlabeled generic data combined with unlabeled data collected from the target parking lot during the first $N$ days of deployment, enabling the incorporation of target-domain information into the self-supervised stage.

Finally, in the third stage, a linear classifier is placed on top of the encoder, and the entire model is fine-tuned in a supervised manner using labeled data from generic parking lots.
This procedure is applied to both the \emph{Strong General Model} and the \emph{Specialized Model}. 
The intuition behind the proposed training pipeline is straightforward.
Generic unlabeled images are first used to learn robust and transferable representations.
These representations are then refined for the parking lot domain using unlabeled parking spot images.
Finally, the resulting domain-adapted features are guided toward the target classes, empty or occupied, through supervised fine-tuning on generic parking lot data.

%% file: 4-experiment_protocol/experiment_protocol.tex
\section{Experimental Setup}
\label{sec:experiment_protocol}

This section describes the datasets, evaluation protocol, and training configuration used in our experiments.

\subsection{Datasets}
\label{subsec:datasets}

Following prior work, we use ImageNet~\cite{imagenet} for pretraining on general-purpose visual data.
For domain-specific data, we employ three parking spot occupancy datasets: PKLot~\cite{pklot}, CNRPark-EXT~\cite{cnrpark}, and PLds~\cite{plds}.
These datasets include images captured under different parking lots, cameras, viewing angles, and weather conditions, as summarized in \cref{tab:datasets}\footnote{New annotations were added and previous annotations were standardized to allow a fair comparison of results.}.

\begin{table}[htbp!]
    \centering
    \caption{Summary of the parking-space classification datasets.}
    
    \vspace{-1.75mm}
    
    \label{tab:datasets}
    \resizebox{\linewidth}{!}{
    \begin{tabular}{lccccc}
      \toprule
      Dataset & Annotations & Days & \makecell{Parking\\Lots} & Angles & \makecell{Weather\\Conditions} \\
      \midrule
      PKLot~\cite{pklot} & 1,199,857 & 100 & 2 & 3 & 3 \\
      CNRPark-EXT~\cite{cnrpark} & 165,513 & 23 & 1 & 9 & 3 \\
      PLds~\cite{plds} & 104,728 & 115 & 1 & 3 & 5 \\
      \bottomrule
    \end{tabular}
    }
\end{table}

The PKLot dataset contains images collected at the Universidade Federal do Paraná and the Pontifícia Universidade Católica do Paraná, both in Brazil.
It comprises approximately 1.2 million annotated cropped parking space images, with an average resolution of $57 \times 59$ pixels.
PKLot includes data from three cameras, UFPR04, UFPR05, and PUCPR, and covers three weather conditions: sunny, rainy, and cloudy.

The CNRPark-EXT dataset consists of approximately 165,000 annotated samples collected at the National Research Council in Italy.
Each cropped image has an average resolution of $96 \times 91$ pixels.
The dataset includes cameras 1 to 9 and the same weather conditions as PKLot.

\begin{figure}[htbp]
    \captionsetup{captionskip=2.5pt}
    \centering    
    \subfloat[PKLot~\cite{pklot}]{\includegraphics[width=.48\linewidth]{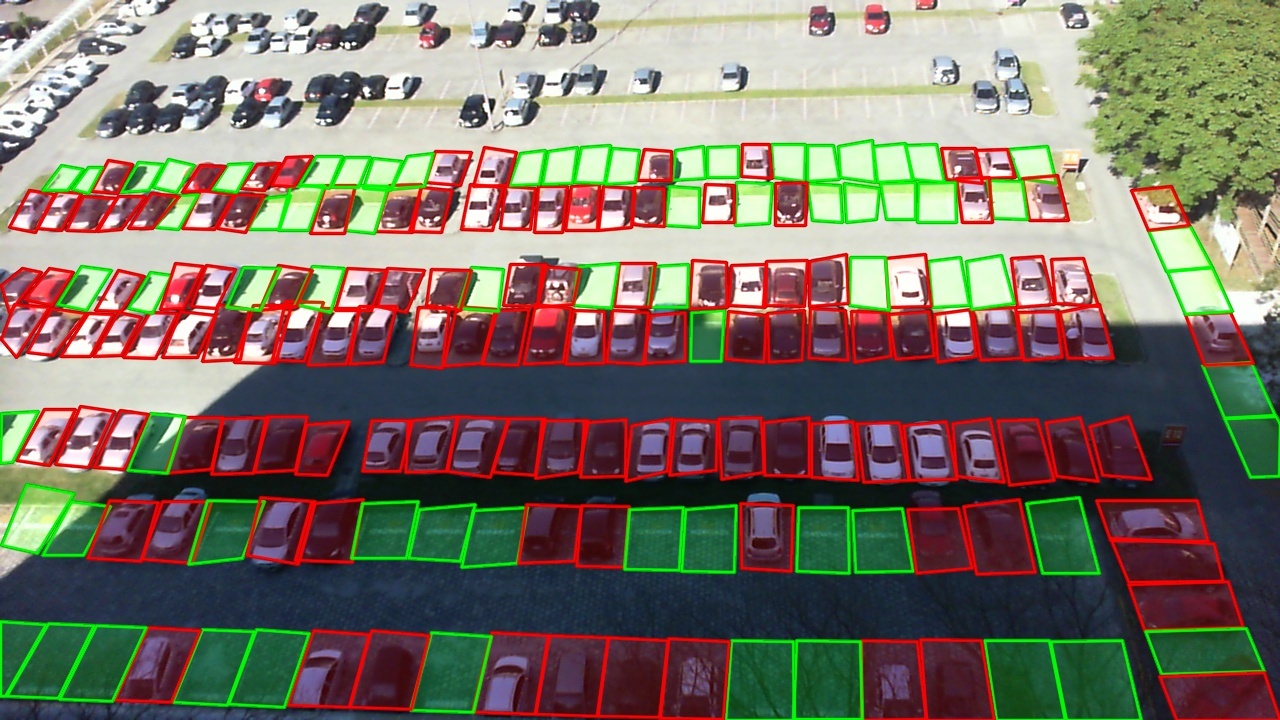}}
    \subfloat[CNRPark-EXT~\cite{cnrpark}]{\includegraphics[trim={0 0 0.5cm 0},clip,width=.48\linewidth]{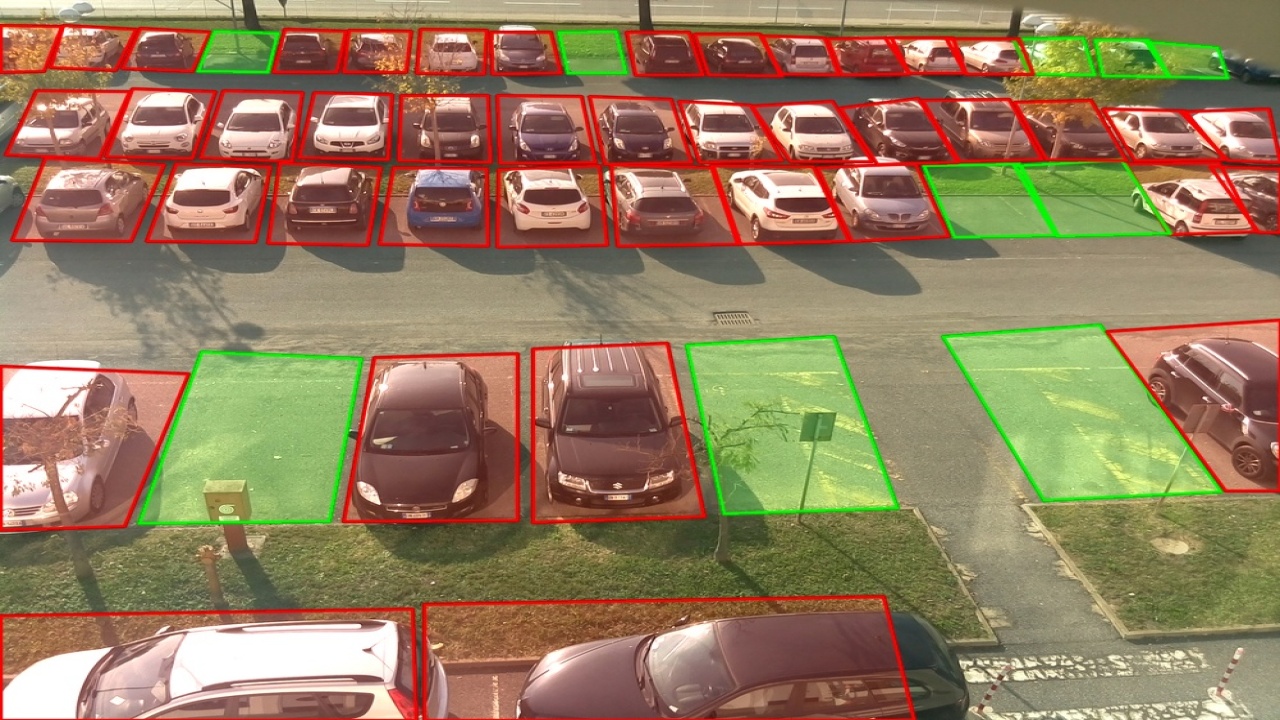}}

    \vspace{-0.75mm}
    
    \subfloat[PLds~\cite{plds}]{\includegraphics[width=.52\linewidth]{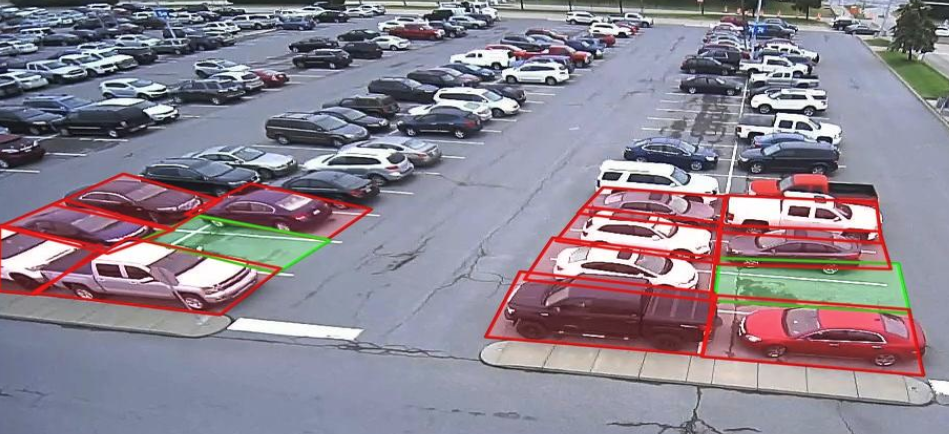}}
    
    \caption{Examples from the parking lot datasets explored in this work.}
    \label{fig:datasets}
\end{figure}

The PLds dataset contains approximately 104,000 labeled samples with an average resolution of $303 \times 108$ pixels, collected at the Pittsburgh International Airport.
It includes five weather conditions, sunny, rainy, cloudy, snowy, and clear night, as well as three cameras\footnote{As suggested in \cite{paulo_review}, we merged vxusd and vmlix.}: isshk, qridr, and vxusd/vmlix.
Example images from each dataset are shown in \cref{fig:datasets}.

To avoid biases, we follow the evaluation protocol suggested in~\cite{paulo_review}, which is also adopted in \cite{cnrpark} and \cite{paulo_distill}.
Specifically, we employ a leave-one-out procedure to assess cross-environment generalization.
Models are trained on two datasets and evaluated on the remaining one.
The evaluation splits are: training on PKLot and CNRPark-EXT and testing on PLds; training on PKLot and PLds and testing on CNRPark-EXT; and training on CNRPark-EXT and PLds and testing on PKLot.

\subsection{Base Model and Hyperparameter Settings}

As in \cite{simclr}, we employ the ResNet-50 architecture~\cite{resnet} as the backbone of the SimCLR framework.
Both the proposed \emph{Strong General Model} and \emph{Specialized Model} follow a training pipeline composed of two self-supervised stages followed by supervised fine-tuning, as illustrated in \cref{fig:deploy_scheme}.
All input images are resized to $224 \times 224$ and normalized using the ImageNet mean and standard deviation.

We initialize the encoder using publicly available SimCLR weights\footnote{\url{https://github.com/google-research/simclr}.}, corresponding to the first self-supervised stage.
In the second stage, self-supervised training is performed on task-specific data.
Following~\cite{medical_image_classification}, we employ the LARS optimizer with a learning rate of $0.3$, a batch size of $512$, a weight decay of $10^{-6}$, NT-Xent temperature of $\tau = 0.5$, and project the representation to a 128-dimensional latent space in the non-linear projection head.
The total number of training steps is set to 80,000, without the use of learning rate scheduling or warm-up.

The number of steps was determined by temporarily splitting each training set into training and validation subsets, where the smaller dataset was used for validation (e.g., when using PKLot + PLds as the training set, PLds served as the validation set). The network was initially trained for 200,000 steps, and the optimal number of steps was selected by monitoring the validation accuracy using a classifier head added on top of the encoder, with the entire model fine-tuned following the protocol described in~\cite{simclr}. Once the optimal number of steps was identified, the model was retrained on the complete dataset (i.e., training + validation). We adopt the same data augmentation strategy as in \cite{simclr}: random resized cropping with a uniform scale from 0.08 to 1.0 and an aspect ratio ranging from 3/4 to 4/3, random horizontal flipping~($p = 0.5$), color jittering~($p = 0.8$), random grayscale conversion~($p = 0.2$), and Gaussian blur~($p = 0.5$).

In the third stage, supervised fine-tuning is conducted on task-specific data.
Following~\cite{medical_image_classification}, we use the SGD optimizer with Nesterov momentum of 0.9, trained for 30,000 steps with a batch size of 256.
For each training and validation split defined in the second stage, we perform a grid search over learning rates $\in [10^{-3.5}, 10^{-3}, 10^{-2.5}, 10^{-2}, 10^{-1.5}, 10^{-1}, 10^{-0.5}]$ and weight decay values $\in [10^{-5}, 10^{-4}, 10^{-3}, 0]$.
No data augmentation is applied during supervised fine-tuning, and both the \emph{Strong General Model} and \emph{Specialized Model} use identical hyperparameter configurations.

We compare our approach against two baselines: a \emph{Supervised Baseline} and a \emph{Self-supervised Baseline}.
In both cases, standard ResNet-50 models are trained following conventional transfer learning via fine-tuning protocols, as described in~\cite{simclr}.
Each baseline includes an initial pretraining phase on ImageNet, performed either in a supervised\footnote{Supervised ResNet-50 weights are available at \url{https://docs.pytorch.org/vision/main/models/generated/torchvision.models.resnet50}.} or self-supervised manner.
This stage is followed by supervised fine-tuning on task-specific data using the same strategy adopted in the proposed method.
For fairness, learning rate and weight decay values are selected through the same grid search procedure.

Finally, following~\cite{paulo_distill}, we define $N = 7$ as the number of days before switching from the \emph{Strong General Model} to the \emph{Specialized Model} in all experiments.

%% file: 5-results/results.tex
\section{Results and Discussion}
\label{sec:results}

The results reported in this section correspond to the average of five runs.
All experiments were conducted on a server equipped with two Intel Xeon Gold 6430 processors, 512 GB of DDR5 DRAM, and an NVIDIA RTX 6000 Ada Generation GPU with 48 GB of VRAM.
To simulate a real-world deployment on an edge device, we also conducted experiments on a Raspberry Pi 5 with 8 GB of~DRAM.

\subsection{Main Results}
\label{subsec:comparison}

\cref{tab:results} presents the results for each leave-one-out split.
As the dataset subsets are relatively well balanced, results are reported in terms of accuracy.
Both the \emph{Self-supervised Baseline} and \emph{Supervised Baseline} achieved strong performance, with overall average accuracy rates of 97.0\% and 97.1\%, respectively.
These results exceed the average accuracy of 91.8\% reported in~\cite{paulo_review} for cross-parking-lot evaluation scenarios and remain consistently high across all leave-one-out splits, which we primarily attribute to the network depth and the large amount of generic parking lot data available during training.

\begin{table}[!htbp]
    \caption{Accuracy (\%) of baselines and proposed methods across multiple parking lot datasets.}

    \centering

    \vspace{-1.75mm}

    \begin{tabular}{@{} ccccc @{}}
    \toprule
    \makecell{Test\\Subset} &
    \makecell{Self-supervi-\\sed Baseline} &
    \makecell{Supervised\\Baseline} &
    \makecell{Strong Gen-\\eral Model\\(ours)} &
    \makecell{Two-stage\\Deployment\\(ours)} \\
    \midrule 
    \multicolumn{5}{c}{\textbf{PKLot}} \\[0.5ex]
    UFPR04 & 98.1 $\pm$ 0.7 & 97.5 $\pm$ 0.8 & 98.6 $\pm$ 0.1 & 98.9 $\pm$ 0.4 \\
    UFPR05 & 96.1 $\pm$ 0.7 & 96.6 $\pm$ 0.4 & 97.3 $\pm$ 0.2 & 97.9 $\pm$ 0.1 \\
    PUCPR & 97.0 $\pm$ 0.7 & 97.3 $\pm$ 0.2 & 97.0 $\pm$ 0.1 & 97.9 $\pm$ 0.1\\
    Average & 97.0 & 97.2 & 97.3 & 98.0 \\
    \midrule \midrule
    \multicolumn{5}{c}{\textbf{CNRPark-EXT}} \\[0.5ex]
    camera1 & 96.5 $\pm$ 0.1 & 95.3 $\pm$ 0.6 & 94.7 $\pm$ 0.8 & 95.1 $\pm$ 0.5 \\
    camera2 & 99.5 $\pm$ 0.1 & 99.2 $\pm$ 0.2 & 98.9 $\pm$ 0.4 & 99.2 $\pm$ 0.3 \\
    camera3 & 98.5 $\pm$ 0.1 & 97.4 $\pm$ 0.4 & 97.9 $\pm$ 0.9 & 98.0 $\pm$ 0.5 \\
    camera4 & 97.7 $\pm$ 0.1 & 98.0 $\pm$ 0.1 & 97.7 $\pm$ 0.4 & 97.8 $\pm$ 0.2 \\
    camera5 & 95.1 $\pm$ 0.2 & 96.3 $\pm$ 0.2 & 97.1 $\pm$ 0.5 & 97.2 $\pm$ 0.4 \\
    camera6 & 96.9 $\pm$ 0.1 & 96.2 $\pm$ 0.3 & 96.3 $\pm$ 0.2 & 96.5 $\pm$ 0.2 \\
    camera7 & 94.6 $\pm$ 0.2 & 94.2 $\pm$ 0.2 & 95.2 $\pm$ 0.1 & 95.5 $\pm$ 0.5 \\
    camera8 & 98.4 $\pm$ 0.1 & 98.0 $\pm$ 0.2 & 97.8 $\pm$ 0.3 & 97.8 $\pm$ 0.5 \\
    camera9 & 94.1 $\pm$ 0.1 & 94.8 $\pm$ 0.4 & 96.2 $\pm$ 0.3 & 96.3 $\pm$ 0.4 \\
    Average & 96.3 & 96.5 & 96.7 & 96.8 \\ 
    \midrule \midrule
    \multicolumn{5}{c}{\textbf{PLds}} \\[0.5ex]
    isshk & 95.8 $\pm$ 0.8 & 95.8 $\pm$ 0.9 & 94.4 $\pm$ 1.1 & 94.4 $\pm$ 1.4 \\
    qridr & 99.2 $\pm$ 0.1 & 98.9 $\pm$ 0.5 & 99.2 $\pm$ 0.2 & 99.1 $\pm$ 0.2 \\
    vxusd/vmlix & 98.0 $\pm$ 0.3 & 98.9 $\pm$ 0.2 & 98.4 $\pm$ 0.6 & 98.1 $\pm$ 0.7 \\
    Average & 97.2 & 97.6 & 96.8 & 96.7 \\ 
    \midrule \midrule
    \textbf{Global Avg.} & 97.0 & 97.1 & 97.2 & 97.8 \\
    \bottomrule
    \end{tabular}
    \label{tab:results}
\end{table}

The \emph{Strong General Model} column reports classification results obtained using only the trained \emph{Strong General Model}, without switching to the Specialized Model after the $N$-th day.
This approach slightly outperformed the baselines, achieving an overall average accuracy of 97.2\%. However, it is observed that the \textit{Strong General Model} struggled on the PLds dataset, yielding results inferior to those of the baselines.
This behavior suggests that the generic features extracted by the encoder may not be sufficiently discriminative for this dataset.
PLds poses additional challenges due to low-angle image captures (see \cref{fig:datasets}), which often result in occlusions.

The proposed \emph{Two-stage Deployment} scheme, in which the \textit{Strong General Model} is deployed during the first $N$ days and the \textit{Specialized Model} thereafter, achieved the best overall performance, with a global average accuracy of 97.8\%.
This strategy improves performance across two datasets compared to the \emph{Strong General Model} alone.
Nevertheless, on the PLds dataset, the \emph{Two-stage Deployment} underperformed the \emph{Supervised Baseline} by 0.7 percentage points.

Overall, the results obtained by both the \emph{Strong General Model} and the \emph{Two-stage Deployment} indicate that self-supervised learning can achieve strong performance for parking spot occupancy classification.
Even without access to unlabeled samples from the target domain, the \emph{Strong General Model} outperformed the baselines on average, and the obtained results are competitive with those reported in previous works.
For instance, \cite{paulo_distill}, which also adopts a two-stage deployment strategy, reports average accuracy values of 95.3\% for the teacher model and 97.0\% for the student model.
Our proposed approach, on the other hand, achieved an average accuracy of 97.3\% in the first $N$ days with the \emph{Strong General Model}, and 97.9\% in the subsequent days using the \emph{Specialized Model}. Similarly, Hochuli et al.~\cite{hochuli} reported an accuracy of 90.1\% when evaluating on the PKLot~dataset.

Finally, we present the results regarding the training and inference times. Training a \emph{Strong General Model} or a \emph{Specialized Model} takes approximately 25 hours on a single GPU. 
When deployed on a Raspberry Pi, classifying a single parking spot takes an average of 0.2282 seconds, of which 0.2140 seconds correspond to inference and the remaining time to pre-processing steps such as image cropping.
Thus, images from a parking lot with 100 visible parking spaces can be processed in under 23 seconds on an edge device.
Although this runtime is approximately 20 times slower than that reported in~\cite{paulo_distill}, it remains acceptable for periodically refreshing parking-space occupancy status. Nevertheless, we note that processing time may become prohibitive when using more severely power-constrained edge~devices.

\subsection{Using Labeled Samples from the Target Parking Lot}

\begin{figure}[htbp]
    \centering
    \hspace{-0.2cm}
    \input{5-results/tikz/label_efficiency}
    \vspace{-0.7cm}
    \caption{Accuracy of the models when labeled samples from the target parking lot are given for training. The labeled samples come from the first \emph{N} days, and evaluation is done in the remaining days.}
    \label{fig:label_efficiency}
    \vspace{-0.22cm}
\end{figure}
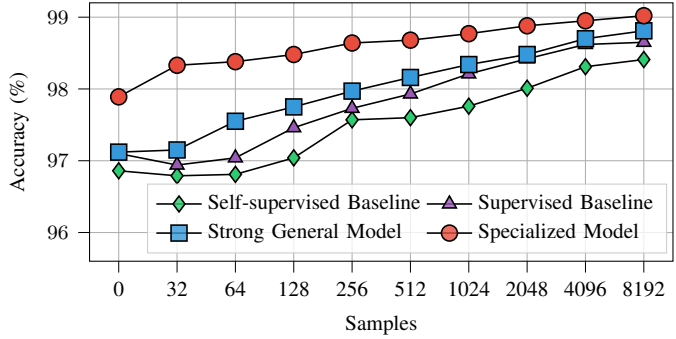

Using a strategy similar to that of \cite{hochuli2022annotStrat}, here we evaluate the label efficiency of the proposed approach when labeled samples from the target parking lot are available.
\cref{fig:label_efficiency} illustrates the relationship between classification accuracy and the number of labeled samples.
The labeled samples are taken randomly from the first $N$ days of deployment, and evaluation is performed considering only the remaining days.
As can be seen, the \emph{Specialized Model} consistently outperforms all competing methods across the entire range of labeled samples. As a comparison, the approach proposed in \cite{hochuli2022annotStrat} achieves an accuracy of 97\%\footnote{This comparison must be considered with caution, since in \cite{hochuli2022annotStrat} only the PKLot dataset was used, and the model has considerably fewer parameters.} when 1,000 samples from the target are given, while our proposed approach reaches an accuracy of~98.8\%.

It is also worth noting that the \emph{Self-supervised Baseline} exhibited a slower rate of improvement and a stronger dependence on labeled data, as its performance with 8,192 samples remains noticeably lower than that of all other models, including the supervised one.

%% file: 5-results/tikz/label_efficiency.tex
\begin{tikzpicture}

\definecolor{darkgray176}{RGB}{176,176,176}
\definecolor{dodgerblue52152219}{RGB}{52,152,219}
\definecolor{lightgray204}{RGB}{204,204,204}
\definecolor{mediumorchid15589182}{RGB}{155,89,182}
\definecolor{mediumseagreen46204113}{RGB}{46,204,113}
\definecolor{tomato2317660}{RGB}{231,76,60}

\begin{axis}[
height=5cm,
width=9.3cm,
legend cell align={left},
legend style={
  fill opacity=0.8,
  draw opacity=1,
  text opacity=1,
  at={(0.99,0.03)},
  anchor=south east,
  draw=lightgray204,
  legend columns=2,
  font=\footnotesize, %
},
label style={font=\footnotesize}, %
tick label style={font=\footnotesize}, %
title style={font=\footnotesize},
tick align=outside,
tick pos=left,
x grid style={darkgray176},
xlabel={Samples},
xmajorgrids,
xmin=-0.5, xmax=9.5,
xtick style={color=black},
xtick={0,1,2,3,4,5,6,7,8,9},
xticklabels={0,32,64,128,256,512,1024,2048,4096,8192},
y grid style={darkgray176},
ylabel={Accuracy (\%)},
ymajorgrids,
ymin=95.6, ymax=99.2,
ytick style={color=black}
]

\addplot [semithick, black, mark=diamond*, mark size=3, mark options={solid,fill=mediumseagreen46204113}]
table {%
0 96.86
1 96.79
2 96.81
3 97.04
4 97.57
5 97.6
6 97.76
7 98.01
8 98.31
9 98.41
};
\addlegendentry{Self-supervised Baseline}

\addplot [semithick, black, mark=triangle*, mark size=3, mark options={solid,fill=mediumorchid15589182}]
table {%
0 97.10
1 96.94
2 97.04
3 97.46
4 97.73
5 97.93
6 98.21
7 98.42
8 98.62
9 98.65
};
\addlegendentry{Supervised Baseline}

\addplot [semithick, black, mark=square*, mark size=3, mark options={solid,fill=dodgerblue52152219}]
table {%
0 97.12
1 97.15
2 97.55
3 97.75
4 97.97
5 98.16
6 98.34
7 98.48
8 98.7
9 98.81
};
\addlegendentry{Strong General Model}

\addplot [semithick, black, mark=*, mark size=3, mark options={solid,fill=tomato2317660}]
table {%
0 97.89
1 98.33
2 98.38
3 98.48
4 98.64
5 98.68
6 98.77
7 98.88
8 98.95
9 99.02
};
\addlegendentry{Specialized Model}
\end{axis}

\end{tikzpicture}

%% file: 6-conclusion/conclusion.tex
\section{Conclusions}
\label{sec:conclusion}

As discussed in \cite{paulo_review}, recent advances in image-based parking lot management systems should prioritize methods that achieve high accuracy without relying on labeled samples from the target environment, while remaining lightweight. In this work, we introduce a self-supervised approach for parking spot occupancy recognition and evaluate it in a cross-dataset setting using SimCLR and three widely used datasets.
The proposed method is based on a training pipeline composed of two self-supervised phases, one using generic data and another using domain-specific data, followed by supervised fine-tuning with generic labeled parking lot images.

The proposed approach results in a \emph{Strong General Model} that achieves an average accuracy of 97.2\%. When sufficient training resources are available (25 GPU hours), a \emph{Specialized Model} can be trained after the $N$-th day of deployment (the 7th day in our experiments) to replace the \emph{Strong General Model}, yielding an average accuracy of 97.8\%.
This \emph{Two-stage Deployment} scheme requires no manual annotation, making the approach both practical and scalable.

Regarding inference cost, classifying a single parking space image takes 0.2282 seconds on a Raspberry Pi~5. This result indicates that the proposed approach is suitable for edge computing when relatively capable edge devices are available. Nevertheless, inference latency may still be prohibitive for more resource-constrained platforms.

Although computational demands during training and deployment remain a challenge, the overall results demonstrate that self-supervised learning is an effective strategy for achieving both cross-environment generalization and domain specialization. Future work will explore alternative self-supervised paradigms, different model architectures, and trade-offs between accuracy and efficiency for deployment on edge~devices.